# Most Language Models can be Poets too: An AI Writing Assistant and Constrained Text Generation Studio

**Allen Roush**     **Sanjay Basu**     **Akshay Moorthy**     **Dmitry Dubovoy**
Oracle Corporation    Oracle Corporation    University of Oregon    University of Oregon
allen.roush@oracle.com, sanjay.basu@oracle.com,
AkshayMoorthy123@gmail.com, ddubovoy@protonmail.com

## Abstract

Despite rapid advancement in the field of Constrained Natural Language Generation, little time has been spent on exploring the potential of language models which have had their vocabularies lexically, semantically, and/or phonetically constrained. We find that most language models generate compelling text even under significant constraints. We present a simple and universally applicable technique for modifying the output of a language model by compositionally applying filter functions to the language models vocabulary before a unit of text is generated. This approach is plug-and-play and requires no modification to the model. To showcase the value of this technique, we present an easy to use AI writing assistant called "Constrained Text Generation Studio" (CTGS). CTGS allows users to generate or choose from text with any combination of a wide variety of constraints, such as banning a particular letter, forcing the generated words to have a certain number of syllables, and/or forcing the words to be partial anagrams of another word. We introduce a novel dataset of prose that omits the letter "e". We show that our method results in strictly superior performance compared to fine-tuning alone on this dataset. We also present a Huggingface "space" web-app presenting this technique called Gadsby. The code is available to the public here: https://github.com/Hellisotherpeople/Constrained-Text-Generation-Studio

## 1 Introduction

Constrained writing is a literary approach in which the writer decides to impose patterns, constraints, or conditions on their text. The most obvious example of this application is within poetry – but many other communities of writers also find imposing constraints on themselves to be enjoyable. We can divide constraints into two types, *soft-constraints* and *hard-constraints*.

Soft constraints are the kind that are fuzzy, e.g. deciding to write in a certain style. Soft constraints are almost exclusively applied at the sequence level, rather than being applied directly on each token. Hard constraints are concrete lexical, semantic, or phonetic requirements about the contents of a token or sequence. In this paper, we are presenting a system that applied token level hard-constraints to large-scale language models.

One notable group who create hard-constrained texts are the *Oulipo* (short for Ouvroir de littérature potentielle; roughly translated as the "workshop of potential literature") writing collective. Oulipo affiliated writers have produced a prolific amount of constrained literature since the 1960s. Oulipos founder has described the writers within the collective as "rats who construct the labyrinth from which they plan to escape".

One does not need to be a rodent to find "recreational linguistics" useful. Any suitor who has pledged their affection in print can attest to how difficult it can be to write good love poetry; and being able to generate rhyming text that also has the lengths of consecutive words matching the digits of pi is sure to swoon all but the most frigid of mathematicians.

Natural Language Generation has advanced at a breakneck pace. As models have scaled up, their

9

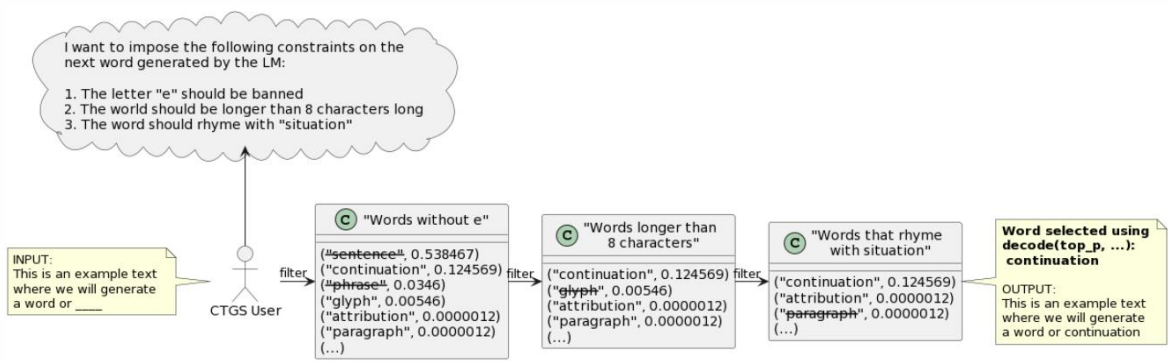

Figure 1: A use-case diagram of the algorithm

performance on a wide variety of tasks has improved. More recent work shows that sufficiently large models such as the "Pathways Language Model" (Chowdhery et al., 2022) unlock new capabilities for common sense reasoning. The probabilistic nature of language models makes their impressive performance particularly intriguing.

Ultimately, all language models involve some form of sampling from their vocabulary of all possible tokens that they could generate. In this paper, we explore the idea of adding arbitrarily compositional lexical, semantic, and/or phonetic filters to the crucial step of a language model sampling from its vocabulary during its decoding phase. Among other things, we observe that language models can remain coherent even with a remarkable amount of filters applied to their vocabulary. We thus find that it is perfectly appropriate to expect coherent output from a model like GPT-2 (Radford et al., 2019), when, for instance, its vocabulary is filtered to ban any word with the letter "e", the letter "a" is forced to appear, and the length of the token must be longer than 3.

In this paper, we introduce two systems which take advantage of this constrained vocabulary technique: An AI writing assistant called Constrained Text Generation Studio (CTGS) and a Huggingface "space" web-app called "Gadsby"[1].

Constrained Text Generation Studio is a GUI tool for recreational linguists, poets, creative writers, and/or researchers to use and study the ability of large-scale language models to recommend relevant text in nearly any situation. After specifying and downloading one of the thousands of language models made available on the Huggingface model hub, users can use CTGS to specify a list of constraints or "filters" that the vocabulary of the language model must pass through before it can be sampled from. After any combination of the filters are specified, users can either use traditional decoding methods to generate tokens from the constrained vocabulary automatically, or they can manually select their continuation from the list of valid tokens. CTGS was created with the idea of being "like Photoshop but for Constrained Text Generation".

Gadsby is a Huggingface hosted webapp which demonstrates the ability for language models to generate coherent text with several different pre-selected combinations of filters. Gadsby was named after one the most famous constrained works of fiction, which is a 270 page book written without the letter E. Gadsby is missing features that CTGS has, including composability of filters, optional human selection of continuations, and text transforms – but it includes filter pre-sets to showcase the robustness of language models to constraints. The most notable of these pre-sets is called "E-Prime"[2], which filters the specified language models vocabulary to avoid any form of the verb "to be".

## 2 Prior Work

We are not the first to explore Constrained Natural Language Generation with Language Models. Probably the closest prior work to our own comes from Pascual et al. (2021). They propose a single plug-and-play semantic filter which shifts the sampling probabilities of a language models vocabulary towards a user defined keyword or set of keywords. CTGS instead offers a rich array of compositional lexical, phonetic, and semantic filters, and it preserves the original language models sampling probabilities with the exception of the filtered out tokens, which are banned.

---

[1] Available here: https://huggingface.co/spaces/Hellisotherpeople/Gadsby

[2] The wikipedia article about this is fascinating: https://en.wikipedia.org/wiki/E-Prime

Swanson et al. (2014) show that language models using Constrained Beam Search can effectively generate text with the constraint of either banning or requiring certain words to appear in a sequence. Notably, the transformers library from Huggingface recently integrated this functionality [3]. Constrained Beam Search is effective for translation and other sequence-to-sequence tasks, but it makes it impossible for the language model to assist humans on a per-token basis. CTGS adopts an optional human-in-the-loop approach where the user can decide which token to choose following the listed constraints at each step, rather than necessarily relying on sampling. Given the inherit creativity required for Constrained Writing, using language models for inspiration rather than blindly generating with them is uniquely helpful for recreational linguists.

Kumar et al. (2021) propose a method for Controlled Text Generation by formulating it as an optimization problem given a list of constraints and using gradient descent to maximize the log probability of the language model as well as the constraint objectives. The constraints that they provide are exclusively sequence level. By contrast, CTGS's filters are at the token level and are correspondingly much more appropriate for Oulipo or Poetry. Their method also requires a potentially lengthy optimization process.

Lu et al. (2021) propose a reinforcement learning based technique for generating sequences with conceptual constraints. This method requires training and is not applicable for hard lexical or phonetic constraints.

Zhang et al. (2020) developed a technique for solving the problem of hard-constraint generation. They propose to pre-train a model by progressively inserting tokens between existing tokens in a parallel manner. They introduce a large scale language model pre-trained this way and which is fine-tuned on hard-constrained tasks called POINTER. Their work only looks at the constraint of requiring certain words to appear in a sequence. Our work explores a wide variety of constraints and requires no training.

Other work related to constrained text generation which explores the potential of global constraint satisfaction at the sequence level comes from Mireshghallah et al. (2022). Surrogate models, such as BertScore, enforce these global constraints. Our writing assistant enforces constraints at the local level, and allows human intervention at any point.

Some intriguing work from the Task Oriented Dialogue community has parallels with our work. Balakrishnan et al. (2019) showcase how constrained decoding can be obtained by controlled modification of the model representation. They find that this technique improves semantic correctness as measured on the weather dataset.

## 3 Implementation Details

In this section, we explore the quirks, caveats, and details of the implementation of our technique within CTGS.

### 3.1 Filters

To enable a filter, a user checks the corresponding box, which will cause a larger group of settings to become visible. These settings are specific to each individual filter. After the relevant settings are specified, the button at the bottom of the settings enables the filter, and a list of filters which are enabled is shown at the top of the filters window.

CTGS at the time of writing includes 21 filters. Many of these filters are lexical, such as constraints which ban or force particular letters. Other filters are distance based, such as the semantic filter, which uses an auxiliary fasttext (Bojanowski, et al., 2016) model to remove language model vocabulary tokens which don't meet or exceed the specified semantic similarity threshold with a user supplied word.

Probably the most interesting of the included filters are phonetic in nature. CTGS includes filters for syllable count, meter, rhyme, and phonetic matching. CTGS achieves this feat by using the the

---
[3] An excellent blog post about this can be found here: https://huggingface.co/blog/constrained-beam-search

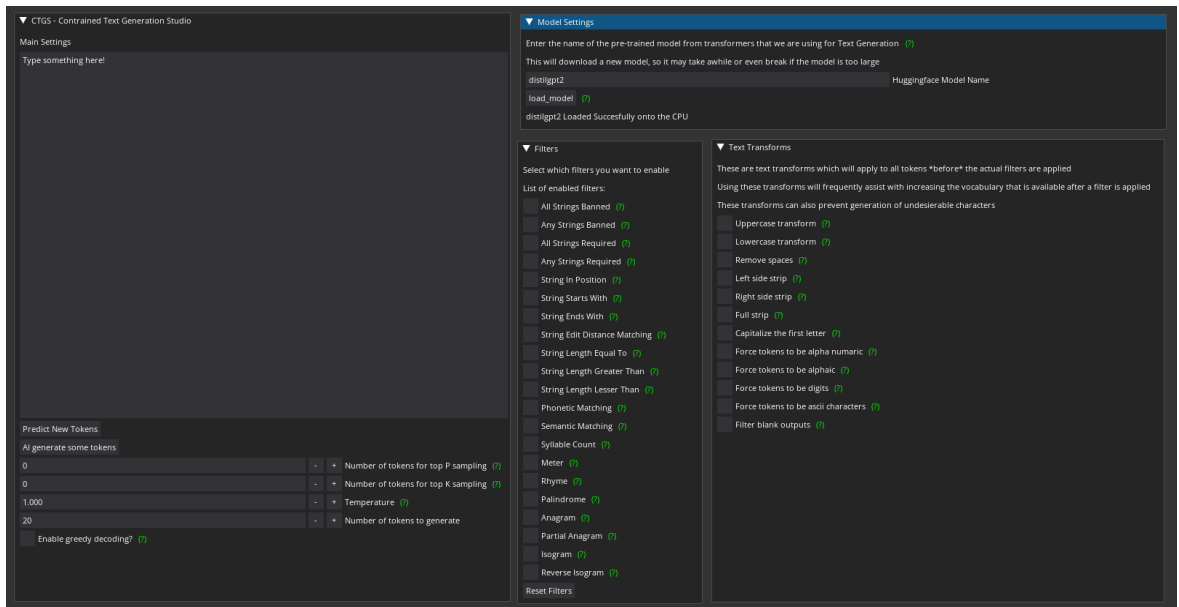

Figure 2: CTGS with the "Distilled-GPT2" (distilgpt2) model loaded. Users can right click within the textbox for a list of all possible continuations matching the currently selected filters

Carnegie Mellon Pronouncing Dictionary (CMUdict)[4]. The "double metaphone" phonetic algorithm is used for direct phonetic matching. These sorts of filters unlock the potential for poetry generation by large-scale language models since the rhyme, syllable, or meter constraints inherit to poetry are directly forced within the language models vocabulary.

### 3.2 Tokenization

Most of the constraints have the additional unpleasant side-effect of subverting the intention and value of subword tokenization schemes. This is because the filters assume that a language model generates its words all in one-step. Subword tokenization became the de facto default for large language models because increasing the vocabulary size of a language model dramatically increases the computational and memory footprint of the model. As the size and sophistication of language models has gone up, their vocabulary sizes have stayed constant[5]. This is frustrating for our technique, which naively assumes that a filter can be applied to a subword – an assumption which is often not true.

Unfortunately, most Language Models don't "signpost" as to whether they are generating a full word or a subword, requiring heuristic techniques to be used if one wanted to construct a "subword aware" filter. Even more startlingly, we observe that language models occasionally generate functionally *the same continuation* with subwords that they could have generated with direct words found within the vocabulary. Many of the filters in CTGS will absolutely cripple a language models ability to generate rare words which would be vectorized into subwords by the language models tokenizer. CTGS in its current form thrives when it is using a language model with a huge vocabulary.

Luckily, modern language models with huge vocabularies exists. One of these is "Transformer-XL", which showcased the value of using a word-tokenizer and an autoregressive architecture for generating coherent text (Dai et al., 2019). Its word-tokenizer doesn't leverage sub words, and thus these models do not succumb into the previously discussed issues. The default pre-trained models that Dai et al made available have a vocabulary size of 267735 tokens. That's a 5.32x increase in size over GPT-3! Unfortunately, one must also incur a significant penalty in memory and compute costs for this privilege.

### 4 Dataset without the "e"

One of the issues that large language models present for constrained writers is that even when heavily fine-tuned on a particular dataset, they

---

[4] Available here: https://github.com/cmusphinx/cmudict

[5] E.g. GPT, GPT-2, and GPT-3 all have a vocab size of 50257 words.

12

frequently *ignore their constraints*. For example, poetry models that were fine-tuned on the works of William Shakespeare frequently stumble and fail to maintain rhyme or meter.[6] We show that language models, which are fine-tuned even on the simple lexical constraint of omitting the letter "e", still occasionally ignore their constraints. In fact, even when these models are overtrained to an absurd degree, complete adherence to these constraints is unlikely.

Such behavior motivates the creation of datasets which include some forms of hard lexical, semantic, or phonetic constraints. By doing so, we can measure how often language models ignore them, and more importantly, we can show that this method of filtering out these tokens before the generation step leads to strictly better performance and eliminates these kinds of errors.

We present a dataset, called "Lipogram-e", which consists of all known complete book-length English works which do not use the letter "e". This dataset includes all of *Gadsby* by Ernest Vincent Wright, all of *A Void* by Georges Perec, and almost all of *Eunoia* by Christian Bok [7]. We name it "Lipogram-e" because a lipogram is a text where the author omits one or more letters from the alphabet.

While it may be possible to produce a dataset without the letter "e" by simply computationally looking through an existing large scale dataset for sentences which match that constraint, doing so would result in jumbled and incoherent training examples, with little relation to each other. By contrast, books and prose written with constraints have clear, coherent narratives. We chose the constraint of banning "e" because it is extremely easy to computationally verify and because there is no potential for error from the filter function.

## 5 Experiment

We design the experiment to measure how often a language model makes constraint-based mistakes on the Lipogram-e dataset. We look at the perplexity and the ignored constraint error rate of GPT-2-medium. We choose GPT-2-medium because of its relatively well-understood fine-tunability. We compare the untrained GPT-2 model to the regularly fine-tuned model, and the over-fine-tuned model. We show that in all instances, applying the constraint to ban the letter "e" from the vocabulary of these models results in both improved perplexity, as well as zero ignored constraint errors.

| Model | Perplexity on test split | Ignored Constraint Error % |
|---|---|---|
| GPT-2 | 237.37 | 28.2 |
| GPT-2 with constraint filter | **211.53** | 0 |
| GPT-2 fine-tuned for 5 epochs | 78.24 | 0.5 |
| GPT-2 fine-tuned for 5 epochs with constraint filter | **77.99** | 0 |
| GPT-2 fine-tuned for 20 epochs | 75.58 | 0.3 |
| GPT-2 fine-tuned for 20 epochs with constraint filter | **75.10** | 0 |

Table 1: Results of the experiment on the Lipogram-e dataset

## 6 Discussion and Observations

Language models that have had their vocabularies filtered act significantly differently from unaltered models. Because the filters remove significant amounts of entries with high probability of being generated, models are more likely to behave undesirably. Some of the undesirable behavior observed included models generating total gibberish, generating repetitive text, generating potentially personally identifying information, generating profanity, and generating computer code. The more tokens which are filtered, and the higher their probability, the more likely it is that models will end up in these degenerate states. We hope that this paper motivates further and more exhaustive analysis of the vocabularies of language models and in particular, what properties they have when altered.

Filtering the vocabularies of language models opens up unique possibilities for adversarial machine learning. Any model which is exposing its full probability distribution before decoding could potentially be "attacked" by a sophisticated actor who has figured out what they "don't want" the

---

[6] An observation that has also been made by others: see here: https://www.gwern.net/GPT-2

[7] *Eunoia* is a work where each chapter only uses one vowel. We omit the chapter that uses the vowel "e"

13

model to generate. This could dramatically reduce the number of generations needed to leak specific information.

Similar techniques for filtering the output of all generative models could be explored in the future. Highly sophisticated text-to-image models like DALL-E from Ramesh et al. (2021) and Stable-Diffusion from Ho and Salimans (2021) might have interesting and unique behavior if pixel based filters that are analogous to our technique can be developed.

It would be extremely interesting to see how this technique will work with large scale language models such as OpenAIs GPT-3 or Huggingfaces BLOOM model. It is likely to make this technique extremely sophisticated, but large scale models frequently are not released to the public and their vocabularies probability distributions are not always exposed to the end user.

# 7 Final Thoughts and Conclusion

In this paper, we introduced the AI constrained writing assistant called CTGS, explained its features and rationale, and mused about its potential use cases. We also introduced a Huggingface hosted webapp which demonstrates the plug-and-play nature of constraining the vocabulary of a language model. We introduced a dataset of English books which do not contain the letter "e" called "Lipogram-e". We showed that our technique results in lower perplexity and zero ignored constraint errors in a variety of circumstances. Finally, we discussed the unique behaviors that models with constraints have.

We also hope to use this paper to serve as a call to action for the language modeling community to not abandon research into word level tokenizers and training models using them. If that's not possible, at least some form of "signposting" should be built into subsequently trained models using potentially a new subword tokenization scheme designed for this purpose. We hope this paper motivates future work on word-level tokenization, and on language models trained with extremely large vocabularies.